\documentclass[11pt,a4paper]{article}
\usepackage[hyperref]{emnlp2017}
\usepackage{times}
\usepackage{latexsym}
\usepackage{url}
\usepackage{helvet}
\usepackage{courier}
\usepackage{amsmath}
\usepackage{graphicx}
\usepackage{fancyhdr}
\usepackage{adjustbox}
\usepackage{multirow}
\usepackage{float}
\usepackage{textcomp}
\usepackage{algorithm,algorithmic}
\usepackage{amssymb,amsthm}
\usepackage{bbm}

\emnlpfinalcopy 

\title{Controlled CNN-based Sequence Labeling for Aspect Extraction}

\author{Lei Shu, Hu Xu, Bing Liu\\
    Department of Computer Science\\
    University of Illinois at Chicago\\
    \{lshu3, hxu48, liub\}@uic.edu
}

\date{}

\begin{document}
\maketitle

\begin{abstract}

One key task of fine-grained sentiment analysis on reviews is to extract aspects or features that users have expressed opinions on. This paper focuses on supervised aspect extraction using a modified CNN called \underline{c}on\underline{tr}o\underline{l}led CNN (Ctrl). The modified CNN has two types of control modules. Through asynchronous parameter updating, it prevents over-fitting and boosts CNN's performance significantly. This model achieves state-of-the-art results on standard aspect extraction datasets. To the best of our knowledge, this is the first paper to apply control modules to aspect extraction.

\end{abstract}

\section{Introduction}
Aspect extraction is an important task in sentiment analysis \cite{HuL2004} and has many applications \cite{Pang2008OMS,Liu2012,Cambria2012}.
It aims to extract opinion targets (or aspects) from opinion text. 
In reviews, aspects are attributes or features of opinion targets. 
For example, from ``\textit{The screen is great}'' in a laptop review, it aims to extract ``screen''. 

Aspect extraction has been performed using supervised and unsupervised approaches. 
Since this work focuses on supervised learning, for existing unsupervised approaches, such as frequent pattern mining \cite{HuL2004,PopescuNE2005}, syntactic rules-based extraction \cite{ZhuangJZ2006,WangBo2008,QiuLBC2011, shu2016lifelong}, topic modeling \cite{MeiLWSZ2007,TitovM2008,Lin2009,Moghaddam2011}, word alignment \cite{KangLiu2013IJCAI} and label propagation \cite{Zhou-wan-xiao:2013:EMNLP}.
Traditionally, supervised approaches \cite{Jakob2010,Mitchell-EtAl:2013:EMNLP} use Conditional Random Fields (CRF) \cite{Lafferty2001conditional}.
Recently, deep networks have also been applied, for example, using
LSTM \cite{williams1989learning,hochreiter1997long,liu2015fine} and
attention mechanism \cite{wang2017coupled,he2017unsupervised} together with manual features \cite{poria2016aspect,wang2016recursive}.
Further, \cite{wang2016recursive,wang2017coupled,li2017deep,li2018aspect} also proposed aspect and opinion terms co-extraction via a deep network.
More recently, a simple CNN model called DE-CNN \cite{xu_acl2018} achieved state-of-the-art performances on aspect extraction by leveraging a double embedding mechanism. Besides using general-purpose embeddings (e.g., GloVe embeddings), DE-CNN also uses domain-specific embeddings \cite{xu2018lifelong} to boost its performance without using any manual feature.

In this paper, we use DE-CNN as a base model. We notice that in traditional CNN model training process, 
all CNN layers are updated together (synchronously) through back-propagation. They easily over-fit the training dataset though validation dataset used for deciding the best parameters' values.
Inspired by a recent work called Deep Adaption Network (DAN) \cite{rosenfeld2017incremental}, we design two kinds of control modules to adjust the input of each CNN layer.
Although DAN is for incremental learning (continually adapt a model for new tasks without losing performance on previous tasks), we observe that by asynchronously updating control modules and CNN layers, it can boost the performance of a single task, too.
The critical point is that we do not train all parameters at the same time. Instead, we optimize CNN layers when we fix control modules' parameters. The control modules work as adding noise on each CNN layer's input. This makes the training little harder and ensures the whole model does not fully fit the training data. After that, we optimize control modules by fixing CNN layers' parameters. Since CNN layers' parameters is optimized on noisy input, in this step, the whole model does not easily over-fit training data as well. In every step (fixing control modules or fixing CNN layers), we track the best validation model and make the next step training start with this best validation model. Once the best validation score does not change after several steps, the whole asynchronous-updating training process completes.

To achieve better efficiency, we propose two kinds of control modules: Embedding Control Module and CNN Control Module. The former is applied after the embedding layer, and the later is applied between two adjacent CNN layers.
Using these control modules and asynchronously updating control modules and CNN layers prevent overfitting.
The experiment results show that this idea is promising. To the best of our knowledge, this is the first paper that incorporates control modules and asynchronously updating.



\section{Related Work}


CNN \cite{lecun1995convolutional,kim2014convolutional,du2017convolutional} is recently adopted for machine translation \cite{gehring2017convolutional}, named entity recognition \cite{kalchbrenner2014convolutional, chiu2015named,ma2016end,strubell2017fast}, sentiment analysis \cite{poria2016aspect, chen2017improving} and aspect extraction \cite{xu_acl2018}. We does not purely use CNN but propose control modules to boost the performance of CNN.

DAN\cite{rosenfeld2017incremental} solves incremental learning problem by (1) training a base CNN network on the initial task, (2) encountering a new task, train on the square linear transformations of the base CNN layer to utilize base CNN network for the new task and also maintain base CNN's performance for the initial task. Residual network\cite{he2016deep} solves gradient vanishing problem on a very deep neural network by providing high-way bridges between CNN layers. We do not solve incremental/transfer learning nor gradient vanishing problems. We do asynchronous parameter update to prevent over-fitting and make the only one task better.

\section{Model}\label{sec:model}

\begin{figure}[!htb]
   \begin{minipage}{0.2\textwidth}
     \centering
     \includegraphics[width=.5\linewidth]{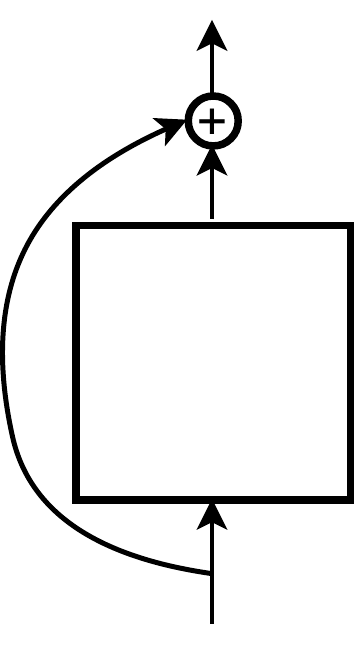}
     \caption{Embedding Control Module}\label{Fig:square}
   \end{minipage}\hfill
   \begin {minipage}{0.2\textwidth}
     \centering
     \includegraphics[width=.5\linewidth]{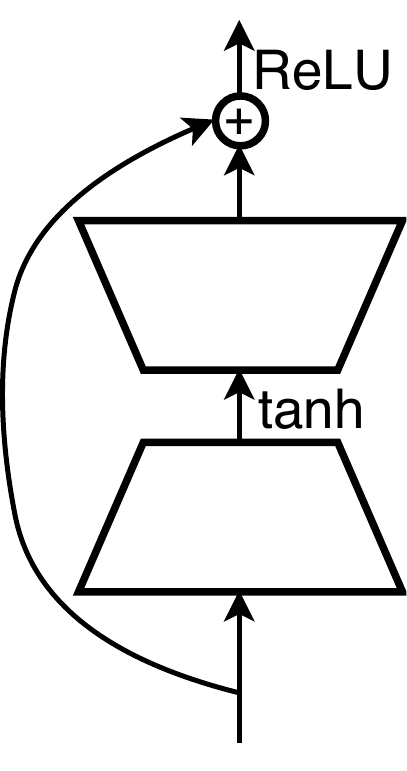}
     \caption{CNN Control Module}\label{Fig:bottleneck}
   \end{minipage}
\end{figure}
\begin{table}[t]
    \centering
    \scalebox{0.82}{
        \begin{tabular}{c|c|c}
        \hline
            {\bf $l$}  &{\bf Type }        &{\bf Parameter Size}   \\\hline
             1 & Emb & (vocab\_size,300) (vocab\_size,100)  \\
             1 & Emb Ctrl & (400, 400)(400,)\\
             2 & CNN & (128, 400, 3)(128,) (128, 400, 5)(128,)  \\
             2 & CNN Ctrl & (256, 128) (128,) (128, 256) (256,)  \\
             3 & CNN & (256, 256, 5) (256,)  \\
             3 & CNN Ctrl & (256, 128) (128,) (128, 256) (256,)  \\
             4 & CNN & (256, 256, 5) (256,)  \\
             4 & CNN Ctrl & (256, 128) (128,) (128, 256) (256,)  \\
             5 & CNN & (256, 256, 5) (256,)  \\
             6 & Linear & (256, 3) (3, )  \\\hline
        \end{tabular}
    }
    \caption{Network layers and parameters, CNN: (filter\_size, in\_dim, kernel\_size), Others: (in\_dim, out\_dim) or (out\_dim)}
    \label{tab:parameter} 
\end{table}
The proposed model is depicted in Tabel \ref{tab:parameter}. It has a double embedding \cite{xu_acl2018} layer (we later use embedding layer for simplicity), multiple CNN layers, multiple control modules, and a fully-connected+softmax layer. 
Note that we keep the architecture of \cite{xu_acl2018} and only add control modules.
We apply control modules after the embedding layer and each CNN layer, except the last CNN layer.

We propose two kinds of control modules.

\textbf{Embedding Control Module} As shown in Figure \ref{Fig:square}, embedding control module adds the input and the transformed input via a square matrix together. The purpose of using this control module is to keep the original embedding and meanwhile slightly adjust the representation of the embedding.

Assume the input is a sequence of word indexes $x=(x_1, \dots, x_n)$.
Let $x^{(1)}$ denote the output from the embedding layer. 
The controlled output from the embedding layer is:
\begin{equation}
\small
z^{(1)}= (w_{\text{emb}}^{(1)} x^{(1)}+b_{\text{emb}}^{(1)}) + x^{(1)},
\end{equation}
where $w_{\text{emb}}^{(1)}$ and $b_{\text{emb}}^{(1)}$ are trainable weights.

\textbf{CNN Control Module} As shown in Figure \ref{Fig:bottleneck}, CNN control module has a bow tie structure. The size of the hidden dimensions is first reduced and later expanded. We use $\text{tanh}(\cdot)$ as the intermediate activation function. To avoid over-fitting, we also apply dropout after this activation function. This bow tie structure can help to strengthen important information from each CNN's output. The expanded output is also added to the output of CNN to keep the original representation with a slight adjustment. Finally, ReLU activation is applied to ensure the output is greater than or equal to 0.

Specifically, let $x^{(l)}$ denote the output of the $(l-1)$-th CNN layer(first layer is embedding layer). The output $z^{(l)}$ of the CNN control module is computed as:

\begin{equation}
\small
\begin{split}
    z^{(l)}= \max\Bigg(0, x^{(l)}+\\\bigg( w_{\text{exp}}^{(l)}
    \text{tanh}(w_{\text{red}}^{(l)} x^{(l)}+b_{\text{red}}^{(l)})
    + b_{\text{exp}}^{(l)}\bigg)
    \Bigg),
\end{split}
\end{equation}

where $w_{\text{exp}}^{(l)}$, $w_{\text{red}}^{(l)}$, $b_{\text{red}}^{(l)}$ and $b_{\text{exp}}^{(l)}$ are trainable weights.


Further, we let $\Theta_{\text{cnn}}$, $\Theta_{\text{ctrl}}$, $\Theta_{\text{fc}}$ denote the trainable parameters in CNN layers, control layers and the final fully connected layer, respectively. 
We define the asynchronous training as follows. At every step, the model is initialized to the previous step's best validation model and save the best validation model during training.

\textbf{Step (1)} fix $\Theta_{\text{ctrl}}$, $\Theta_{\text{fc}}$, and tune on $\Theta_{\text{cnn}}$. 

\textbf{Step (2)} fix $\Theta_{\text{cnn}}$, and tune on $\Theta_{\text{ctrl}}$, $\Theta_{\text{fc}}$.

Repeat step (1) and step (2) until the best validation score does not change after several steps.
In this way, CNN layers are trained when control modules are frozen, and the control modules are trained when the CNN layers are frozen. 

For better comparing with state-of-the-art method DE-CNN~\cite{xu_acl2018}, we keep the embedding and all CNN layers the same as DE-CNN. DE-CNN has a double embedding layer, 4 CNN layers, a fully-connected layer shared across all positions of the words, and a softmax layer over the labeling space $\mathcal{Y}=\{B(-Aspect), I(-Aspect), O(ther)\}$ for each position of inputs. For the first CNN layer, two different filter sizes are employed.
For the rest 3 CNN layers, only one filter size is used. 
We apply dropout after the embedding layer and each ReLU activation.
As the reason indicated by \cite{xu_acl2018}, the double embedding layer is frozen since the training data for aspect extraction is usually small.
The embedding control module lies between the embedding layer and the first CNN layer. Three CNN control modules lie between any two adjacent CNN layers. Details are also in Table \ref{tab:parameter}.

\section{Experiment}
\begin{figure}[htp]
\includegraphics[clip,width=\columnwidth]{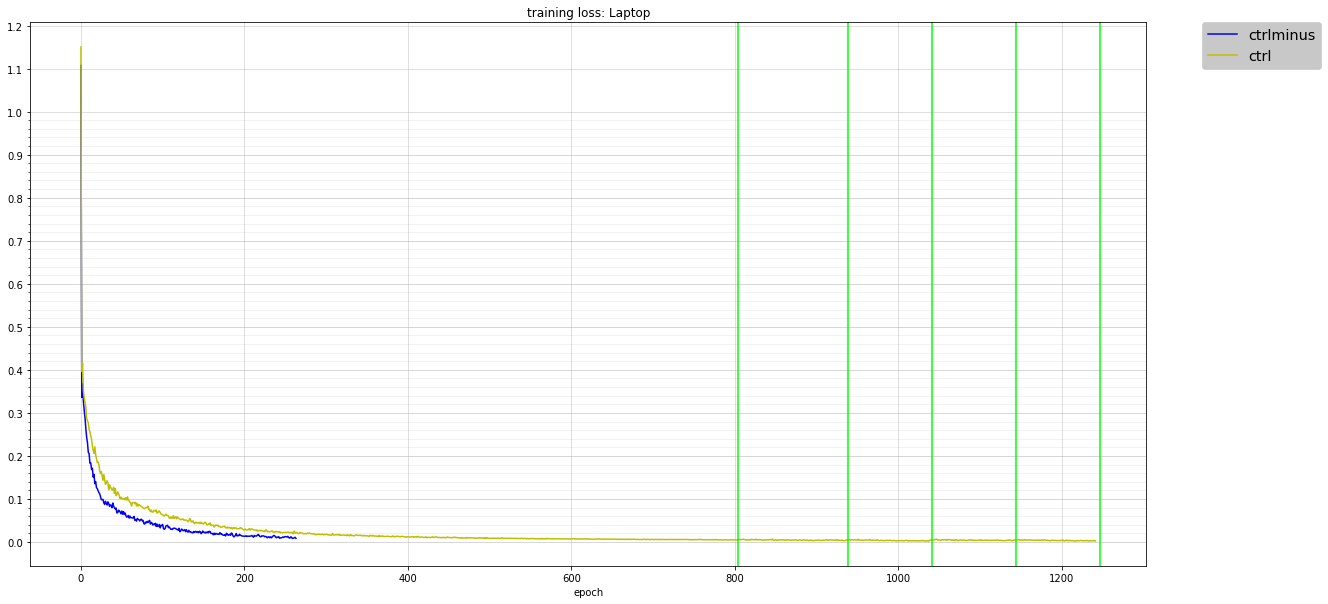}
\includegraphics[clip,width=\columnwidth]{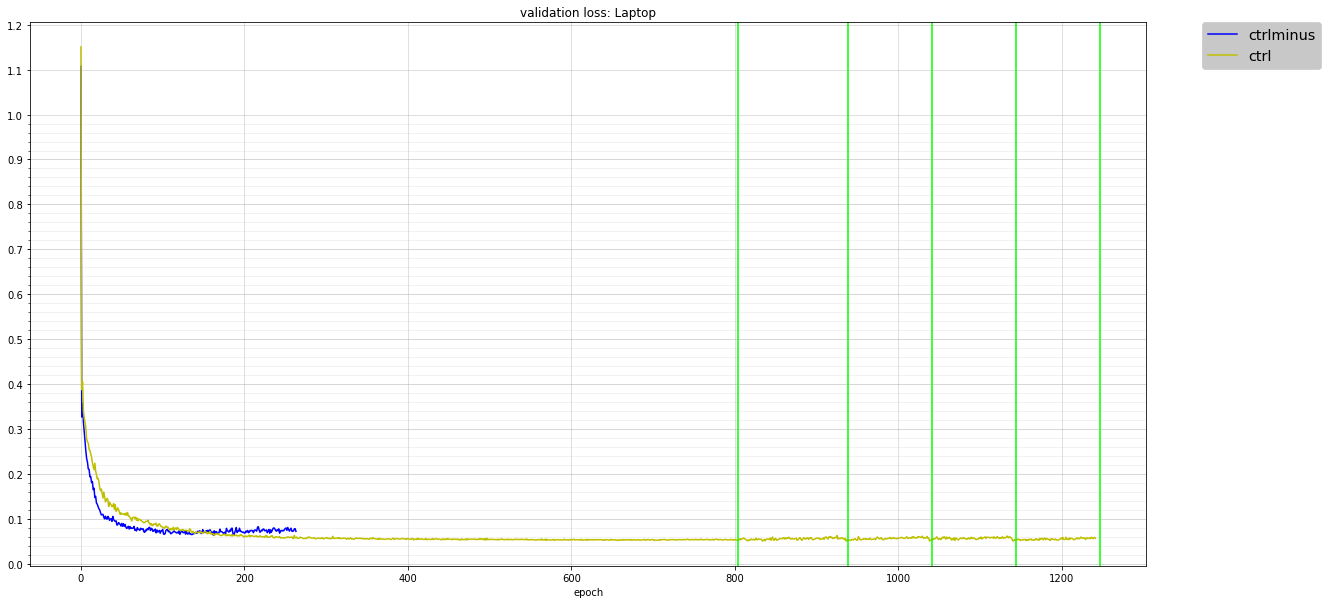}
\includegraphics[clip,width=\columnwidth]{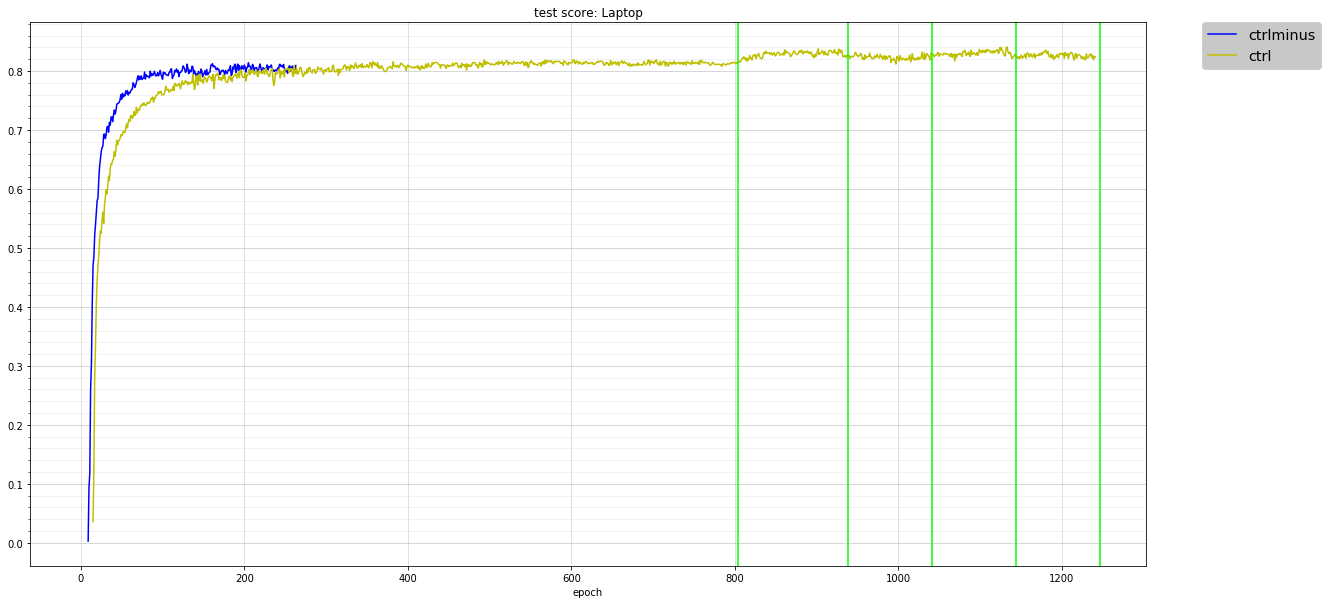}
\caption{They are Ctrl-'s and Ctrl's training loss, validation loss and test score during training on Laptop dataset. Green vertical lines indicate the end of steps.}
\label{fig:laptop}
\end{figure}



We conduct experiments on two benchmark datasets from SemEval challenges \cite{pontiki2014SemEval,pontiki2016semeval}, as shown in Table \ref{tab:dataset}. 
The first dataset is in the \textit{laptop} domain from SemEval-2014 Task 4.
The second dataset is in the \textit{restaurant} domain from SemEval-2016 Task 5.
We use NLTK\footnote{http://www.nltk.org/} to tokenize each sentence. 
For double embedding, the general-purpose embeddings are from the glove.840B.300d embeddings \cite{pennington2014glove}. 
The domain-specific embeddings are obtained from DE-CNN~\cite{xu_acl2018}\footnote{https://github.com/howardhsu/}. We hold out 150 examples from training data as validation data to decide the hyper-parameters.
The dropout rate is 0.55. For asynchronous updating, we use Adam optimizer \cite{kingma2014adam}. The learning rate of Step (1) is 0.00005, and that of Step (2) is 0.0001. This is because CNN training tends to be unstable and Step (1) trains CNN layers that contain the majority of parameters of the network.
\begin{table}[t]
    \centering
    \scalebox{0.85}{
        \begin{tabular}{c|c|c}
        \hline
            {Datasets}  &{Training Set}        &{Testing Set}  \\
                               &{Sent./Asp.} &{Sent./Asp.}  \\\hline
            SemEval-14 Laptop  &3045/2358              &800/654\\\hline
            SemEval-16 Restaurant&2000/1743            &676/622\\\hline
        \end{tabular}
    }
    \caption{Datasets Statistics: number of sentences(Sent.) and number of aspects(Asp.)}
    \label{tab:dataset} 
\end{table}

\begin{table}[t]
    \centering
    \scalebox{0.75}{
        \begin{tabular}{c||c|c}
        \hline
        \hline
        {} &{Laptop Dataset }  &{Restaurant Dataset }  \\\hline
        CRF         &74.01      &69.56  \\
        IHS\_RD     &74.55      &-      \\
        NLANGP      &-          &72.34  \\
        WDEmb       &75.16      &-      \\
        LSTM        &75.71      &70.35  \\
        BiLSTM-CNN-CRF&77.8   & 72.5\\
        RNCRF       &78.42      &-      \\
        CMLA        &77.80      &-      \\
        MIN         &77.58      &73.44  \\
        THA \& STN &79.52      &73.61  \\
        BERT        &77.19   &71.52 \\
        \hline\hline
        DE-CNN      &81.59      &74.37\\
        \hline\hline
        DAN- -    &78.28   & 70.43  \\
        DAN-        &76.68        &72.94 \\
        DAN         &80.24      &73.35\\
        \hline
        Ctrl- -     &79.47      &71.15\\
        Ctrl-    &81.66  & 73.77  \\
        Ctrl &\textbf{82.73}*     &\textbf{75.64}*\\
        \hline
        \end{tabular}
    }
    \caption{Comparison results in F$_1$ score: results are averaged scores of 5 runs. *indicates the result is statistically significant at the level of 0.01.}
    \label{tab:result} 
\end{table}

\subsection{Compared Methods}
We perform a comparison of Ctrl with two groups of baselines.
The results of the first group are non-CNN based methods.
\textbf{CRF} is conditional random fields. 
\textbf{IHS\_RD} \cite{chernyshevich2014ihs} and \textbf{NLANGP} \cite{toh2016nlangp} are the best systems from the original challenges \cite{pontiki2014SemEval,pontiki2016semeval}.
\textbf{WDEmb} \cite{yin2016unsupervised} is enhanced CRF with multiple embeddings. 
\textbf{LSTM} \cite{liu2015fine,li2018aspect} is a BiLSTM implementation. \textbf{BiLSTM-CNN-CRF} \cite{reimers2017reporting} is the state-of-the-art named entity recognition system. \textbf{BERT} \cite{devlin2018bert} fine-tunes pre-trained language model on aspect extraction tasks.
The following methods use multi-task learning and opinion lexicon or human annotation are adopted for opinion supervision:
\textbf{RNCRF} \cite{wang2016recursive} is a recursive neural network and CRF jointed model for aspect and opinion terms co-extraction. \textbf{CMLA} \cite{wang2017coupled} solves the co-extraction through a multi-layer coupled-attention network. \textbf{MIN} \cite{li2017deep} solves co-extraction, and discriminate sentimental/non-sentimental sentences. \textbf{THA \& STN} \cite{li2018aspect} uses opinion summary and aspect  history to improve prediction. 

The second group is a CNN-based method.
\textbf{DE-CNN} \cite{xu_acl2018} is a pure CNN-based sequence labeling model which utilizes double embedding. This is the base model that Ctrl is adapted from. We use this baseline to show the improvements from Ctrl.
The remaining baselines use DE-CNN as the basic network and add an extra intermediate layer between layers in the basic network.
\textbf{DAN}~\cite{rosenfeld2017incremental} adopts linear transformation as control modules for a incremental learning method on image classification. 
\textbf{DAN- -} tunes on all fully connected layers given frozen random-value CNN layers.
\textbf{DAN-} optimizes all parameters in fully connected layers and CNN layers together. \textbf{DAN} utilizes asynchronous training process between all fully connected layers and CNN layers.
\textbf{Ctrl- -} gives random-value CNN layers (un-trainable), tunes on control modules and fully connected layers.
\textbf{Ctrl-} synchronously updates the control modules, CNN layers, and fully connected layer.
\textbf{Ctrl} asynchronously updates parameters. These are variations of our model.


\subsection{Results and Analysis}
From Table \ref{tab:result}, we can see that our model Ctrl performs the best. The variations of Ctrl always out-perform that of DAN. It shows that a purely linear transformation is unable to produce noise and prevent over-fitting. Ctrl - -'s result shows the adaptive ability of the control modules. 
Ctrl - updates all parameters in the overall network synchronously, but under-performs DE-CNN though it has control modules. The reason is that in synchronous updating, control modules just make the overall network deeper. As in Figure \ref{fig:laptop}, the first plot shows that Ctrl- and Ctrl can reach a similar training loss level and Ctrl- is faster. They have the same learning rate. It means that fixed control modules make the training harder. In the second plot, Ctrl- 's validation loss decreases and then increases. This is an apparent over-fitting signal. But, Ctrl's validation loss tends flat even after several-steps training. From the last test-score plot, we can see that Ctrl has similar testing performance as Ctrl- in the first step training. In Ctrl's second step training (between the first and second green lines), the test score continues improving. The results and plots show that
through asynchronous updating, control modules can prevent over-fitting and improve CNN performance.

\section{Conclusion}
We propose to add two kinds of control modules for CNN-based aspect extraction model. Through asynchronous update, our model Ctrl outperforms state-of-the-art methods significantly.
\bibliography{aspect}

\begin{thebibliography}{}
\expandafter\ifx\csname natexlab\endcsname\relax\def\natexlab#1{#1}\fi

\bibitem[{Cambria and Hussain(2012)}]{Cambria2012}
Erik Cambria and Amir Hussain. 2012.
\newblock {\em Sentic Computing Techniques, Tools, and Applications 2nd
  Edition\/}.
\newblock Springer.

\bibitem[{Chen et~al.(2017)Chen, Xu, He, and Wang}]{chen2017improving}
Tao Chen, Ruifeng Xu, Yulan He, and Xuan Wang. 2017.
\newblock Improving sentiment analysis via sentence type classification using
  bilstm-crf and cnn.
\newblock {\em Expert Systems with Applications\/} 72:221--230.

\bibitem[{Chernyshevich(2014)}]{chernyshevich2014ihs}
Maryna Chernyshevich. 2014.
\newblock Ihs r\&d belarus: Cross-domain extraction of product features using
  crf.
\newblock In {\em Proceedings of the 8th International Workshop on Semantic
  Evaluation (SemEval 2014)\/}. pages 309--313.

\bibitem[{Chiu and Nichols(2015)}]{chiu2015named}
Jason~PC Chiu and Eric Nichols. 2015.
\newblock Named entity recognition with bidirectional lstm-cnns.
\newblock {\em arXiv preprint arXiv:1511.08308\/} .

\bibitem[{Devlin et~al.(2018)Devlin, Chang, Lee, and
  Toutanova}]{devlin2018bert}
Jacob Devlin, Ming-Wei Chang, Kenton Lee, and Kristina Toutanova. 2018.
\newblock Bert: Pre-training of deep bidirectional transformers for language
  understanding.
\newblock {\em arXiv preprint arXiv:1810.04805\/} .

\bibitem[{Du et~al.(2017)Du, Gui, Xu, and He}]{du2017convolutional}
Jiachen Du, Lin Gui, Ruifeng Xu, and Yulan He. 2017.
\newblock A convolutional attention model for text classification.
\newblock In {\em National CCF Conference on Natural Language Processing and
  Chinese Computing\/}. Springer, pages 183--195.

\bibitem[{Gehring et~al.(2017)Gehring, Auli, Grangier, Yarats, and
  Dauphin}]{gehring2017convolutional}
Jonas Gehring, Michael Auli, David Grangier, Denis Yarats, and Yann~N Dauphin.
  2017.
\newblock Convolutional sequence to sequence learning.
\newblock {\em arXiv preprint arXiv:1705.03122\/} .

\bibitem[{He et~al.(2016)He, Zhang, Ren, and Sun}]{he2016deep}
Kaiming He, Xiangyu Zhang, Shaoqing Ren, and Jian Sun. 2016.
\newblock Deep residual learning for image recognition.
\newblock In {\em Proceedings of the IEEE conference on computer vision and
  pattern recognition\/}. pages 770--778.

\bibitem[{He et~al.(2017)He, Lee, Ng, and Dahlmeier}]{he2017unsupervised}
Ruidan He, Wee~Sun Lee, Hwee~Tou Ng, and Daniel Dahlmeier. 2017.
\newblock An unsupervised neural attention model for aspect extraction.
\newblock In {\em Proceedings of the 55th Annual Meeting of the Association for
  Computational Linguistics (Volume 1: Long Papers)\/}. volume~1, pages
  388--397.

\bibitem[{Hochreiter and Schmidhuber(1997)}]{hochreiter1997long}
Sepp Hochreiter and J{\"u}rgen Schmidhuber. 1997.
\newblock Long short-term memory.
\newblock {\em Neural computation\/} 9(8):1735--1780.

\bibitem[{Hu and Liu(2004)}]{HuL2004}
Minqing Hu and Bing Liu. 2004.
\newblock Mining and summarizing customer reviews.
\newblock In {\em KDD '04\/}. pages 168--177.

\bibitem[{Jakob and Gurevych(2010)}]{Jakob2010}
Niklas Jakob and Iryna Gurevych. 2010.
\newblock Extracting opinion targets in a single- and cross-domain setting with
  conditional random fields.
\newblock In {\em EMNLP '10\/}. pages 1035--1045.

\bibitem[{Kalchbrenner et~al.(2014)Kalchbrenner, Grefenstette, and
  Blunsom}]{kalchbrenner2014convolutional}
Nal Kalchbrenner, Edward Grefenstette, and Phil Blunsom. 2014.
\newblock A convolutional neural network for modelling sentences.
\newblock {\em arXiv preprint arXiv:1404.2188\/} .

\bibitem[{Kim(2014)}]{kim2014convolutional}
Yoon Kim. 2014.
\newblock Convolutional neural networks for sentence classification.
\newblock {\em arXiv preprint arXiv:1408.5882\/} .

\bibitem[{Kingma and Ba(2014)}]{kingma2014adam}
Diederik~P Kingma and Jimmy Ba. 2014.
\newblock Adam: A method for stochastic optimization.
\newblock {\em arXiv preprint arXiv:1412.6980\/} .

\bibitem[{Lafferty et~al.(2001)Lafferty, McCallum, and
  Pereira}]{Lafferty2001conditional}
John Lafferty, Andrew McCallum, and Fernando~CN Pereira. 2001.
\newblock Conditional random fields: Probabilistic models for segmenting and
  labeling sequence data.
\newblock In {\em ICML '01\/}. pages 282--289.

\bibitem[{LeCun et~al.(1995)LeCun, Bengio et~al.}]{lecun1995convolutional}
Yann LeCun, Yoshua Bengio, et~al. 1995.
\newblock Convolutional networks for images, speech, and time series.
\newblock {\em The handbook of brain theory and neural networks\/}
  3361(10):1995.

\bibitem[{Li et~al.(2018)Li, Bing, Li, Lam, and Yang}]{li2018aspect}
Xin Li, Lidong Bing, Piji Li, Wai Lam, and Zhimou Yang. 2018.
\newblock Aspect term extraction with history attention and selective
  transformation.
\newblock {\em arXiv preprint arXiv:1805.00760\/} .

\bibitem[{Li and Lam(2017)}]{li2017deep}
Xin Li and Wai Lam. 2017.
\newblock Deep multi-task learning for aspect term extraction with memory
  interaction.
\newblock In {\em Proceedings of the 2017 Conference on Empirical Methods in
  Natural Language Processing\/}. pages 2886--2892.

\bibitem[{Lin and He(2009)}]{Lin2009}
Chenghua Lin and Yulan He. 2009.
\newblock Joint sentiment/topic model for sentiment analysis.
\newblock In {\em CIKM '09\/}. pages 375--384.

\bibitem[{Liu(2012)}]{Liu2012}
Bing Liu. 2012.
\newblock {\em Sentiment Analysis and Opinion Mining\/}.
\newblock Morgan {\&} Claypool Publishers.

\bibitem[{Liu et~al.(2013)Liu, Xu, Liu, and Zhao}]{KangLiu2013IJCAI}
Kang Liu, Liheng Xu, Yang Liu, and Jun Zhao. 2013.
\newblock Opinion target extraction using partially-supervised word alignment
  model.
\newblock In {\em IJCAI '13\/}. pages 2134--2140.

\bibitem[{Liu et~al.(2015)Liu, Joty, and Meng}]{liu2015fine}
Pengfei Liu, Shafiq Joty, and Helen Meng. 2015.
\newblock Fine-grained opinion mining with recurrent neural networks and word
  embeddings.
\newblock In {\em Proceedings of the 2015 Conference on Empirical Methods in
  Natural Language Processing\/}. pages 1433--1443.

\bibitem[{Ma and Hovy(2016)}]{ma2016end}
Xuezhe Ma and Eduard Hovy. 2016.
\newblock End-to-end sequence labeling via bi-directional lstm-cnns-crf.
\newblock {\em arXiv preprint arXiv:1603.01354\/} .

\bibitem[{Mei et~al.(2007)Mei, Ling, Wondra, Su, and Zhai}]{MeiLWSZ2007}
Qiaozhu Mei, Xu~Ling, Matthew Wondra, Hang Su, and ChengXiang Zhai. 2007.
\newblock Topic sentiment mixture: Modeling facets and opinions in weblogs.
\newblock In {\em WWW '07\/}. pages 171--180.

\bibitem[{Mitchell et~al.(2013)Mitchell, Aguilar, Wilson, and
  Van~Durme}]{Mitchell-EtAl:2013:EMNLP}
Margaret Mitchell, Jacqui Aguilar, Theresa Wilson, and Benjamin Van~Durme.
  2013.
\newblock Open domain targeted sentiment.
\newblock In {\em ACL '13\/}. pages 1643--1654.

\bibitem[{Moghaddam and Ester(2011)}]{Moghaddam2011}
Samaneh Moghaddam and Martin Ester. 2011.
\newblock {ILDA}: interdependent lda model for learning latent aspects and
  their ratings from online product reviews.
\newblock In {\em SIGIR '11\/}. pages 665--674.

\bibitem[{Pang and Lee(2008)}]{Pang2008OMS}
Bo~Pang and Lillian Lee. 2008.
\newblock Opinion mining and sentiment analysis.
\newblock {\em Found. Trends Inf. Retr.\/} 2:1--135.

\bibitem[{Pennington et~al.(2014)Pennington, Socher, and
  Manning}]{pennington2014glove}
Jeffrey Pennington, Richard Socher, and Christopher Manning. 2014.
\newblock Glove: Global vectors for word representation.
\newblock In {\em Proceedings of the 2014 conference on empirical methods in
  natural language processing (EMNLP)\/}. pages 1532--1543.

\bibitem[{Pontiki et~al.(2016)Pontiki, Galanis, Papageorgiou, Androutsopoulos,
  Manandhar, Mohammad, Al-Ayyoub, Zhao, Qin, De~Clercq
  et~al.}]{pontiki2016semeval}
Maria Pontiki, Dimitris Galanis, Haris Papageorgiou, Ion Androutsopoulos,
  Suresh Manandhar, AL-Smadi Mohammad, Mahmoud Al-Ayyoub, Yanyan Zhao, Bing
  Qin, Orph{\'e}e De~Clercq, et~al. 2016.
\newblock Semeval-2016 task 5: Aspect based sentiment analysis.
\newblock In {\em Proceedings of the 10th international workshop on semantic
  evaluation (SemEval-2016)\/}. pages 19--30.

\bibitem[{Pontiki et~al.(2014)Pontiki, Galanis, Pavlopoulos, Papageorgiou,
  Androutsopoulos, and Manandhar}]{pontiki2014SemEval}
Maria Pontiki, Dimitris Galanis, John Pavlopoulos, Harris Papageorgiou, Ion
  Androutsopoulos, and Suresh Manandhar. 2014.
\newblock \href{http://www.aclweb.org/anthology/S14-2004}{Semeval-2014 task 4:
  Aspect based sentiment analysis}.
\newblock In {\em Proceedings of the 8th International Workshop on Semantic
  Evaluation (SemEval 2014)\/}. Association for Computational Linguistics and
  Dublin City University, Dublin, Ireland, pages 27--35.
\newblock
  \href{http://www.aclweb.org/anthology/S14-2004}{http://www.aclweb.org/anthology/S14-2004}.

\bibitem[{Popescu and Etzioni(2005)}]{PopescuNE2005}
Ana-Maria Popescu and Oren Etzioni. 2005.
\newblock Extracting product features and opinions from reviews.
\newblock In {\em HLT-EMNLP '05\/}. pages 339--346.

\bibitem[{Poria et~al.(2016)Poria, Cambria, and Gelbukh}]{poria2016aspect}
Soujanya Poria, Erik Cambria, and Alexander Gelbukh. 2016.
\newblock Aspect extraction for opinion mining with a deep convolutional neural
  network.
\newblock {\em Knowledge-Based Systems\/} 108:42--49.

\bibitem[{Qiu et~al.(2011)Qiu, Liu, Bu, and Chen}]{QiuLBC2011}
Guang Qiu, Bing Liu, Jiajun Bu, and Chun Chen. 2011.
\newblock Opinion word expansion and target extraction through double
  propagation.
\newblock {\em Computational Linguistics\/} 37(1):9--27.

\bibitem[{Reimers and Gurevych(2017)}]{reimers2017reporting}
Nils Reimers and Iryna Gurevych. 2017.
\newblock Reporting score distributions makes a difference: Performance study
  of lstm-networks for sequence tagging.
\newblock In {\em Proceedings of the 2017 Conference on Empirical Methods in
  Natural Language Processing\/}. pages 338--348.

\bibitem[{Rosenfeld and Tsotsos(2017)}]{rosenfeld2017incremental}
Amir Rosenfeld and John~K Tsotsos. 2017.
\newblock Incremental learning through deep adaptation.
\newblock {\em arXiv preprint arXiv:1705.04228\/} .

\bibitem[{Shu et~al.(2016)Shu, Liu, Xu, and Kim}]{shu2016lifelong}
Lei Shu, Bing Liu, Hu~Xu, and Annice Kim. 2016.
\newblock Lifelong-rl: Lifelong relaxation labeling for separating entities and
  aspects in opinion targets.
\newblock In {\em Proceedings of the 2016 Conference on Empirical Methods in
  Natural Language Processing\/}. pages 225--235.

\bibitem[{Strubell et~al.(2017)Strubell, Verga, Belanger, and
  McCallum}]{strubell2017fast}
Emma Strubell, Patrick Verga, David Belanger, and Andrew McCallum. 2017.
\newblock Fast and accurate entity recognition with iterated dilated
  convolutions.
\newblock In {\em Proceedings of the 2017 Conference on Empirical Methods in
  Natural Language Processing\/}. pages 2670--2680.

\bibitem[{Titov and McDonald(2008)}]{TitovM2008}
Ivan Titov and Ryan McDonald. 2008.
\newblock A joint model of text and aspect ratings for sentiment summarization.
\newblock In {\em ACL '08: HLT\/}. pages 308--316.

\bibitem[{Toh and Su(2016)}]{toh2016nlangp}
Zhiqiang Toh and Jian Su. 2016.
\newblock Nlangp at semeval-2016 task 5: Improving aspect based sentiment
  analysis using neural network features.
\newblock In {\em Proceedings of the 10th international workshop on semantic
  evaluation (SemEval-2016)\/}. pages 282--288.

\bibitem[{Wang and Wang(2008)}]{WangBo2008}
Bo~Wang and Houfeng Wang. 2008.
\newblock Bootstrapping both product features and opinion words from chinese
  customer reviews with cross-inducing.
\newblock In {\em IJCNLP '08\/}. pages 289--295.

\bibitem[{Wang et~al.(2016)Wang, Pan, Dahlmeier, and Xiao}]{wang2016recursive}
Wenya Wang, Sinno~Jialin Pan, Daniel Dahlmeier, and Xiaokui Xiao. 2016.
\newblock Recursive neural conditional random fields for aspect-based sentiment
  analysis.
\newblock {\em arXiv preprint arXiv:1603.06679\/} .

\bibitem[{Wang et~al.(2017)Wang, Pan, Dahlmeier, and Xiao}]{wang2017coupled}
Wenya Wang, Sinno~Jialin Pan, Daniel Dahlmeier, and Xiaokui Xiao. 2017.
\newblock Coupled multi-layer attentions for co-extraction of aspect and
  opinion terms.
\newblock In {\em AAAI\/}. pages 3316--3322.

\bibitem[{Williams and Zipser(1989)}]{williams1989learning}
Ronald~J Williams and David Zipser. 1989.
\newblock A learning algorithm for continually running fully recurrent neural
  networks.
\newblock {\em Neural computation\/} 1(2):270--280.

\bibitem[{Xu et~al.(2018{\natexlab{a}})Xu, Liu, Shu, and Yu}]{xu_acl2018}
Hu~Xu, Bing Liu, Lei Shu, and Philip~S. Yu. 2018{\natexlab{a}}.
\newblock Double embeddings and cnn-based sequence labeling for aspect
  extraction.
\newblock In {\em ACL\/}.

\bibitem[{Xu et~al.(2018{\natexlab{b}})Xu, Liu, Shu, and Yu}]{xu2018lifelong}
Hu~Xu, Bing Liu, Lei Shu, and Philip~S Yu. 2018{\natexlab{b}}.
\newblock Lifelong domain word embedding via meta-learning.
\newblock In {\em Proceedings of the 27th International Joint Conference on
  Artificial Intelligence\/}. AAAI Press, pages 4510--4516.

\bibitem[{Yin et~al.(2016)Yin, Wei, Dong, Xu, Zhang, and
  Zhou}]{yin2016unsupervised}
Yichun Yin, Furu Wei, Li~Dong, Kaimeng Xu, Ming Zhang, and Ming Zhou. 2016.
\newblock Unsupervised word and dependency path embeddings for aspect term
  extraction.
\newblock {\em arXiv preprint arXiv:1605.07843\/} .

\bibitem[{Zhou et~al.(2013)Zhou, Wan, and Xiao}]{Zhou-wan-xiao:2013:EMNLP}
Xinjie Zhou, Xiaojun Wan, and Jianguo Xiao. 2013.
\newblock Collective opinion target extraction in {Chinese} microblogs.
\newblock In {\em EMNLP '13\/}. pages 1840--1850.

\bibitem[{Zhuang et~al.(2006)Zhuang, Jing, and Zhu}]{ZhuangJZ2006}
Li~Zhuang, Feng Jing, and Xiao-Yan Zhu. 2006.
\newblock Movie review mining and summarization.
\newblock In {\em CIKM '06\/}. pages 43--50.

\end{thebibliography}
\bibliographystyle{emnlp_natbib}

\end{document}